\RequirePackage{etoolbox}
\csdef{input@path}%
{%
 {sty/},% class file
}
\documentclass{elsevierbook}
\usepackage{tcolorbox}
\usepackage{color, colortbl}
\definecolor{LightGray}{rgb}{0.78, 0.79, 0.85}
\definecolor{LightCyan}{rgb}{0.,0.5,0.71}
\usepackage{natbib}
\begin{document}

\Frontmatter

\Mainmatter
  \begin{frontmatter}

\chapter{Machine Learning Robustness: A Primer}\label{chap1}
\subchapter{Houssem Ben Braiek and Foutse Khomh\\
\small{\{houssem.ben-braiek, foutse.khomh\}@polymtl.ca}
}

% \minitoc

\begin{abstract}
This chapter explores the foundational concept of robustness in Machine Learning (ML) and its integral role in establishing trustworthiness in Artificial Intelligence (AI) systems. The discussion begins with a detailed definition of robustness, portraying it as the ability of ML models to maintain stable performance across varied and unexpected environmental conditions. ML robustness is dissected through several lenses: its complementarity with generalizability; its status as a requirement for trustworthy AI; its adversarial vs non-adversarial aspects; its quantitative metrics; and its indicators such as reproducibility and explainability. The chapter delves into the factors that impede robustness, such as data bias, model complexity, and the pitfalls of underspecified ML pipelines. It surveys key techniques for robustness assessment from a broad perspective, including adversarial attacks, encompassing both digital and physical realms. It covers non-adversarial data shifts and nuances of Deep Learning (DL) software testing methodologies. The discussion progresses to explore amelioration strategies for bolstering robustness, starting with data-centric approaches like debiasing and augmentation. Further examination includes a variety of model-centric methods such as transfer learning, adversarial training, and randomized smoothing. Lastly, post-training methods are discussed, including ensemble techniques, pruning, and model repairs, emerging as cost-effective strategies to make models more resilient against the unpredictable. This chapter underscores the ongoing challenges and limitations in estimating and achieving ML robustness by existing approaches. It offers insights and directions for future research on this crucial concept, as a prerequisite for trustworthy AI systems.
\end{abstract}

\begin{keywords}
\kwd{Machine Learning}
\kwd{Deep Learning}
\kwd{Robust AI}
\kwd{Trustworthy AI}
\kwd{Adversarial Robustness}
\kwd{Non-Adversarial Robustness}
\kwd{Model Verification}
\kwd{DL Software Testing}
\kwd{Robust Training}
\kwd{Robustness Assurance}
\end{keywords}

\end{frontmatter}

\section{Definition}\label{definition}
In general, robustness is a predicate that applies to a single entity. For instance, we might consider a sensor robust if it is resilient to disturbances from the environment. In more detail, robustness refers to the ability of a system, model, or entity to maintain stable and reliable performance across a broad spectrum of conditions, variations, or challenges, demonstrating resilience and adaptability in the face of uncertainties or unexpected changes. Hence, we define below the Machine Learning (ML) Model robustness based on the general definition of robustness in ML outlined by Freiesleben and Grote in~\cite{freiesleben2023beyond}. 

\begin{tcolorbox}[colback=gray!8,colframe=gray!40!black]
\textit{ML Model robustness denotes the capacity of a model to \textit{sustain stable predictive performance} in the face of \textit{variations and changes in the input data}.}
\end{tcolorbox}

According to this definition, ML Model robustness is an epistemic concept that presupposes the generalizability of the model’s inductive bias on the in-distribution data, and extends further to evaluate the model's stability and resilience in real-world deployment scenarios. The following Table presents concrete illustrations of how performance degradation and data changes manifest in real-world scenarios.
\[
\begin{array}{|l|}
\hline
\textbf{Examples of variations and changes in the input data:} \\
\hline
\text{-- Variations in input features or object recognition patterns that challenge} \\
\text{the inductive bias learned by the model from the training data.} \\
\text{-- Production data distribution shifts due to naturally occurring distortions,} \\
\text{such as lighting conditions or other environmental factors.} \\
\text{-- Malicious input alterations that are deliberately introduced by an attacker} \\
\text{to fool the model or even steer its prediction in a desired direction.} \\
\text{-- Gradual data drift resulting from external factors, such as evolution in} \\
\text{social behavior and economic conditions.} \\
\hline
\textbf{Examples of model flaws and threats to stable predictive performance:} \\
\hline
\text{-- Exploitation of irrelevant patterns and spurious correlations that will not} \\
\text{hold up in production settings.} \\
\text{-- Difficulty in adapting to edge-case scenarios that are often underrepresented} \\
\text{by training samples.} \\
\text{-- Susceptibility to adversarial attacks and data poisonings that target the} \\
\text{vulnerabilities of overparametrized modern ML models.} \\
\text{-- Inability of the model to generalize well to gradually-drifted data, leading} \\
\text{to concept drift as its learned concepts become obsolete or less representative} \\
\text{of the current data distribution.} \\
\hline
\end{array}
\]

Nevertheless, leaving the range of input data changes unspecified makes it hard to assess the robustness of ML model in practice. We should define the data changes against which the model would be tested. Even if the naturally occurring data distribution shifts are often unanticipated, we usually come up with data distortions that can serve as a proxy for unforeseen data shifts and help us compare the robustness of different ML models. Furthermore, the objective of sustaining stable predictive performance is vague. It is often sufficient for the model to maintain its performance to a certain degree (i.e., the tolerance level) against unexpected changes in input data. This level of tolerance depends on the application context and the assurances needed. For example, the target tolerance for an ML model designed to support clinical decision-making is considerably lower than for an ML model that has been designed to detect spam emails~\cite{freiesleben2023beyond}.\\
Based on these considerations, we can refine further the ML model robustness as follows:
\begin{tcolorbox}[colback=gray!8,colframe=gray!40!black]
\label{rob_def}
\textit{When deployed in a production environment, an ML model is considered robust if variations of input data, as specified by a domain of potential changes, do not degrade the model's predictive performance below the permitted tolerance level.}
\end{tcolorbox}

\subsection{Robustness complements (i.i.d.) generalizability}
In supervised learning, models are commonly estimated via empirical risk minimization (ERM)~\cite{goodfellow2016deep}, a principle that considers minimizing the average loss on observed samples of data, as an empirical estimate of the true risk, i.e., the expected true loss for the entire input distribution. ERM assumes that training and test data are identically and independently distributed, known as closed-world assumption or simply the i.i.d. assumption. The i.i.d generalization refers to the ability of a trained model to deal with novel data inputs, but drawn from the same or close distribution as the training set, called in-distribution (ID) data. The i.i.d. generalization ensures stable predictive performance under static environmental conditions, but it provides no guidance on how to handle out-of-distribution data (OOD)~\cite{yang2021generalized}. In contrast, robustness focuses on capturing the level of predictive performance maintained by the trained model in dynamic environment settings, where input data constantly changes. Robustness could be of little concern if the model fails to i.i.d. generalize well. In order to achieve robustness, we consider i.i.d. generalization to be a necessary but not sufficient condition. For instance, ML models might fail to i.i.d. generalize due to unreliable inductive bias (shortcut learning) or under-fitting (trained with too little/biased data). These models will most likely also perform poorly when input data distributions change.

The ascent of DL has initiated a new era in AI, empowering models to tackle open-world learning challenges in domains like face recognition~\cite{guo2016ms} and autonomous driving~\cite{bojarski2016end}. The term ``generalization" has expanded beyond denoting the model's performance strictly in an i.i.d. setting, encompassing its ability to cope with OOD situations, hereafter referred to as OOD generalization. Although OOD generalization seems to fit the definition of robustness~\ref{rob_def}, it actually refers to the overall predictive performance of an ML model beyond its in-data distribution. It lacks a proper definition of its scope and success criteria since there is no indication of which data distributions the model should generalize to, or how strict the original predictive performance should be maintained. In contrast, ML model robustness is an inherently causal concept since it concerns two causally related entities: the level of predictive performance and the domain of input data changes, so its assessment, as a model's property, requires detailed specification of both entities. Freiesleben and Grote~\cite{freiesleben2023beyond} provide further insight into the complementarity between robustness and generalizability.

\subsection{Robustness is a requirement of Trustworthy AI}
The adoption of ML, especially deep neural networks, has been largely promoted as a result of its impressive performance in terms of accuracy. Meanwhile, a variety of challenges outside of accuracy expectations have emerged, such as malicious attacks against ML-powered systems and misuses of ML that could be harmful. As a result, the Artificial Intelligence (AI) trustworthiness standards~\cite{liu2022trustworthy, li2023trustworthy} have been established to outline representative requirements for current AI systems.  These encompass six critical facets: (i) Safety \& Robustness, (ii) Nondiscrimination \& Fairness, (iii) Transparency \& Explainability, (iv) Privacy, (v) Accountability \& Auditability, and (vi) Environmental Well-being. The complex interplay between these aspects is vital in fostering trustworthy real-world AI systems. For instance, maintaining data privacy might interfere with the desire to explain the system output in detail.

In this sense, robustness is an integral part of AI trustworthiness, while interacting and combining with the other aspects. In fact, model robustness is a cornerstone of safety because robust AI systems are able to deal with unexpected inputs and perturbations without compromising their functionality. It is especially crucial to ensure the appropriate level of model robustness in safety-critical applications, where erroneous behaviors and failures can have catastrophic consequences. The trustworthiness of AI systems requires the integration of two pivotal elements along with robustness assurance: reliable quantification of uncertainty and effective out-of-distribution detection capabilities.\\
\textbf{Uncertainty quantification}~\cite{abdar2021review} refers to methodologies for evaluating uncertainties associated with predictions made by a ML model. It involves assessing the confidence levels or lack thereof in the model's predictions, taking into account factors such as data variance and model error. The quantified uncertainties combined with predictions enable more informed decision-making with AI. 
\begin{tcolorbox}[colback=gray!8,colframe=gray!40!black]
\textbf{Aleatoric uncertainty} stems from inherent randomness in the data, such as noise, and is considered non-reducible. It reflects the unpredictability associated with the inherent variability within observations.
\end{tcolorbox}
\begin{tcolorbox}[colback=gray!8,colframe=gray!40!black]
\textbf{Epistemic uncertainty} arises from limitations within the model's bias and the training data used. It is considered reducible, as it can be mitigated through improvements in modeling strategy (e.g. ensembles), training procedures, or new data collection.
\end{tcolorbox}
In simpler terms, the uncertainty component provides the AI systems with a way to “know what they do not know”. Its main contribution is that uncertain predictions can be ignored from the decision-making flow, avoiding risks in real-world applications. Estimating uncertainties significantly benefits the model robustness analysis by providing an essential means to gauge how individual predictions respond to changes in their data points. This aids in delimiting the domain of input changes on which an ML model is expected to be robust.\\
\textbf{OOD detection}~\cite{yang2021generalized} involves identifying OOD instances at test time that differ significantly from the in-data distribution and might result in mispredictions. It also serves the same purpose of recognizing the boundaries within which the model's patterns are applicable, avoiding the use of its predictions when such restrictions are violated. OOD detection strengthens the reliability of uncertainty quantification by filtering out unusual inputs, on which the uncertainties, as any statistically-inferred estimates, are unlikely to be dependable. Therefore, OOD detection also contributes to further refinement of the data changes domain during the model robustness analysis, and its deployment in safety-critical systems ensures effective operation of robust ML models.

\subsection{Adversarial vs Non-adversarial robustness}
Adversarial robustness is concerned with changes in data distribution that are induced by adversaries to deceive or mislead the ML model. Adversarial distribution shifts can be described as deliberate alterations to original data distribution. The alterations initially focused on introducing well-crafted but imperceptible noises in the data. As an example, a human-imperceptible noise can be applied to medical images to falsify the diagnostic by misleading an ML model into labeling moles as malignant skin tumors~\cite{finlayson2019adversarial}. Then, the input alterations include intelligently-designed changes that are perceptible and can be applied in physical real-world environments. For example, adding sunglasses to a face image is different from slightly distorting the image pixels when evaluating the adversarial robustness of a face detection model. Therefore, adversarial robustness consists of enhancing the model's resilience against these subtle, non-random data distribution shifts without compromising its predictive performance on genuine data.\\
Non-adversarial robustness studies the model's ability to maintain its performance across data distribution shifts arising from naturally-occurring distortions or synthetic data variations that represent conditions more likely to occur in the real world. For instance, a natural shift in images of traffic signs, collected in an area where it rarely snows, can be images of the same sign under severe snowing conditions~\cite{gojic2023non}. Alternatively, a partial discoloration of the traffic sign image, i.e., a region replaced by white pixels, may mimic the effect of snow or other neutral obstructions. Natural data shifts often result from changing environmental conditions that lead to mismatch between the deployment and the training distribution~\cite{xie2019multi}. The changes can be both temporal (i.e., changes in patient demographics, genetic evolution of contagious viruses and their adaptation to host populations) or non-temporal (i.e., variations between patient groups, differences in imaging technologies used by diagnostic devices). The non-adversarial robustness of the ML model ensures reliability across various real-world scenarios, including natural noise, or changes in inputs that might happen organically without malicious intent.
\subsection{Robustness Measurement}
We now present several robustness scoring metrics developed by researchers to evaluate the stability of models’ performance when data changes~\cite{drenkow2021systematic}. These metrics are designed with the assumption of having two datasets: one clean, sampled from ID data, and another perturbed, sampled from shifted or even OOD data. Specifically, shifted data involves significant variations and unusual scenarios within the same domain as the ID data, while OOD data represents inputs from domains that are fundamentally different from the ID data.

Robustness score~\cite{laugros2019adversarial} measures the accuracy loss due to the $\phi$ perturbation. For a given model, m, we denote $A_{\text {clean }}$ as the accuracy of the model, m, on the original (clean) test dataset, whereas $A_\phi$ is the accuracy of the model, $f$, on the test set modified with a $\phi$ perturbation. Data modifications may stem from natural perturbations collected from deployment environments, or they may result from applying an adversarial perturbation to the samples of the original test set. In the same form as~\cite{laugros2019adversarial}, we formulate the robustness score of $f$ to a perturbation in the data inputs, noted $\phi$, with the expression: 

\[ R_f^\phi=\frac{A_\phi}{A_{\text {clean }}} \]

The more the robustness score of a model is close to one, the more it is robust to the considered perturbation. To measure the robustness score of a neural network, $f$, to a set of perturbations $S$, we can use:

\[ R_f^S=\sum_{\phi \in S} R_N^\phi \]

Hendrycks and Dietterich~\cite{hendrycks2019benchmarking} define robustness as average-case performance over a set of corruptions, leading to the definition of mean Corruption Error (mCE) and relative Corruption Error (rCE).
\[
\begin{aligned}
m C E & =\sum_{c \in \mathcal{C}} \sum_{s=1}^5 \frac{E_s^c}{E_{\text {AlexNet }, s}^c} \\
r C E & =\sum_{c \in \mathcal{C}} \sum_{s=1}^5 \frac{E_s^c-E_{\text {clean }}}{E_{\text {AlexNet }, s}^c-E_{\text {AlexNet }, \text { clean }}}
\end{aligned}
\]
where $E$ is the classifier’s error rate, $\mathcal{C}$ is a set of corruptions, and $s$ is the severity of the corruption (refer to Section~\ref{non_adv_shifts} for examples). The AlexNet model~\cite{krizhevsky2012imagenet} served as a common reference point among models. 

Effective and relative robustness~\cite{taori2020measuring} consider how the performance on natural distribution shifts related to the performance on the original test set. The metrics are defined as follows:
\[
\begin{aligned}
\rho(f) & =a c c_2(f)-\beta \cdot a c c_1(f) ,\\
\tau\left(f^{\prime}\right) & =a c c_2\left(f^{\prime}\right)-a c c_2(f)
\end{aligned}
\]
where $f$ is the model under test, $a c c_1$, $a c c_2$ are the accuracy on the original and shifted datasets respectively, $\beta$ is a log-linear fit to the baseline accuracy of a large set of independent models on the original (clean) test set, and $f^{\prime}$ represents the model resulting from a robustness improvement. The effective robustness captures how well a specific model does beyond what is expected given a group of models in general. The relative robustness quantifies the effect of a robustness amelioration strategy on the accuracy under a distribution shift. Overall, a robust model should obtain both positive effective and relative robustness scores. 
\subsection{Robustness Indicators: Reproducibility and Explainability}
So far, robustness has been considered as a quality predicate for an ML model that should be carefully specified, evaluated, and sometimes certified before the deployment of the model in production. Nevertheless, the ML model is the result of a ML engineering process, thus some desirable properties in regards to this process should be satisfied in order to prepare the grounds to reach the desired level of robustness.

Reproducibility controls the randomness of an ML workflow, making it resilient to slight variations in data samples, i.e, there are high chances to converge to the same conclusions when run on two samples from the same underlying distribution~\cite{albertoni2023reproducibility}. For researchers in all scientific fields, including ML, reproducibility is essential to achieve the same results with the same data and algorithms. The absence of reproducibility may lead to claim gains from changing one parameter while the real source of improvement is a hidden source of randomness. Reproducibility is also a strong stability property since it ensures a high probability of replicating results when the datasets are drawn from the same distribution. Consequently, achieving such property on a given learning problem ensures less bias in the estimated model, $f$, since achieving such property confirms that the difference between the empirical risk of $f$ obtained on the training samples and the (true) risk of $f$ is marginal, which implies that M is more likely independent of the training set. For instance, \cite{underspecificationMLmodel} constructs an ensemble of predictors from a given model by perturbing components of the ML pipeline like random seeds for initialization, and retraining the model several times. As a result, output differences between predictors against stress tests can be used as a conservative indicator of the ML pipeline credibility, i.e., its ability to produce a robust model. 

Explainability addresses how an AI model makes decisions~\cite{carvalho2019machine}. Being aware of the reasoning behind predictions can be a fundamental factor that determines the trust in an ML model. DL models are complex and known to share “blackbox” nature, which raises concerns about their deployment in real-world applications despite their better performance compared to intrinsically-interpretable models such as decision trees and linear models. Researchers have developed post-hoc explainers that identify a complex model’s behavior by analyzing its inputs, intermediate results, and outputs. In complex models, a nonlinear regression function or nonlinear classification boundary is learned. A representative category of post-host explainers approximates these non-linear decision functions either globally or locally by using an explainable ML model, i.e., an explainer, such as a linear model~\cite{carvalho2019machine} and rules~\cite{guidotti2018local}. For deep convolutional neural networks or transformers, the inspection of intermediate features is a widely used means of explaining model behavior~\cite{tu2020select,yang2018hotpotqa}. As an example from image classification, saliency maps~\cite{simonyan2013deep} indicate how much influence each input feature (i.e. pixel) has on predicting a specific class. With color coding for their value ranges, they can visually show the regions of interest in the image that have the greatest impact on the classification result. Arun et al.~\cite{arun2021assessing} leverages a variety of saliency map techniques to interpret DNNs trained on medical imaging. In terms of robustness, our aim is sustainable predictive performance under data changes, especially, unexpected distribution shifts. Post-hoc explainers can provide other predictive performance indicators on the model’s behaviors in addition to the accuracy of predictions. Indeed, inconsistent post-hoc explanations for novel inputs when compared with ID samples may indicate erroneous model behaviors due to changes in the data distribution. For instance, saliency maps can generate adversarial noise~\cite{JSM} affecting the most significant regions of interest, and ML testing methods~\cite{braiek2020testing} can be used to find edge cases that lead the model to mistakenly rely on irrelevant regions. They can be seen as ways to capture the distribution shifts from the lens of the model itself by estimating how its behavior (e.g., as captured by post-hoc explanations) differ from the ``normal" behavior observed on the ID samples.  
\section{Challenges}\label{challenges}
\subsection{Data Bias: Train-Serving Skew}
The goal of classical ML and statistical learning theory is to achieve stable performance in an i.i.d. environment, the training data and the unseen data come from the same distribution~\cite{bishop2006pattern}. However, the i.i.d. assumption is not satisfied in most cases because the construction of training datasets with high probability to represent the true data distribution (the inputs in a production environment) is extremely difficult~\cite{sugiyama2012machine}. Real-world data is multi-faceted and virtually infinite, whereas training datasets are finite and constrained by the resources available during the dataset preparation. Data bias can occur in many ways, misrepresenting training datasets for real-world applications. This leads to a common failure known as the train-serving skew~\cite{breck2017ml}: models that perform well in development but poorly in deployment.

There are two main categories of data bias in ML applications: erroneous bias and discriminatory. \textbf{Erroneous bias} can be viewed as a systematic error caused by faulty assumptions. For instance, due to selection bias~\cite{marlin2012collaborative} or sampling bias~\cite{mehrabi2021survey}, the chosen training samples may not be able to represent the real data distribution. 
\begin{tcolorbox}[colback=gray!8,colframe=gray!40!black]
\textbf{Selection bias} refers to a systematic error introduced in the data collection process where certain individuals are more likely to be selected than others, leading to a biased representation of the population.
\end{tcolorbox}
\begin{tcolorbox}[colback=gray!8,colframe=gray!40!black]
\textbf{Sampling bias} occurs when the method used to retrieve data favors certain individuals or groups over others, resulting in a skewed or unrepresentative sample compared to the population of interest.
\end{tcolorbox}
\textbf{Measurement bias} happens when the device used to measure the signal has systematic value distortion that tends to skew the data in a particular direction that prevents the generalization of other data collected by other devices. Discriminatory bias reflects a model’s unfair behaviors toward a certain group or individual, such as producing discriminatory content or performing less well for a specific population or group~\cite{shah2019predictive}. The model is likely to inherit this discrimination bias from the training datasets, leading to undesirable performance on minority groups. 

\textbf{Long-tailed distributions} represent scenarios where certain classes or categories are significantly more prevalent than others, mirroring real-world data collection practices. These distributions are influenced by natural events, and not necessarily introduced by biased data collection. Nonetheless, ML algorithms fail to handle them properly as they are statistically optimized to perform well on common inputs (i.e. the head of the distribution) but struggle where examples are sparse (the tail). The tail often comprises the largest proportion of possible inputs, which makes their inclusion a laborious iterative procedure. Hence, ML models optimized on ID data following long-tail distributions are ineffective in dealing with edge cases because ERM relies on average cost minimization to learn statistically a model that is often biased towards the most common cases. \textbf{Edge cases} denote inputs lying at the extremes or boundaries of the expected range, often residing in the sparse tail of the data distribution. These cases are less frequently encountered in typical usage and common circumstances, but failing to consider them may result in unexpected behaviors and vulnerabilities in the ML model. In safety-critical domains such as self-driving cars and medical diagnosis, failure to handle edge cases properly may have catastrophic consequences, underscoring the importance of gathering and addressing them comprehensively~\cite{brooks2017edge}. Therefore, a robust ML workflow must address the challenges associated with long-tailed distribution modeling, especially considering edge cases that can have detrimental impacts in real-world use cases. 
\subsection{Model Complexity: A Double-Edge Sword}
Conventionally, it is assumed that the use of models with increasing capacity will systematically result in overfitting the training data. Hence, the capacity of a model is usually controlled either by limiting the size of the model (number of parameters) or via various explicit or implicit regularizations, such as early stopping~\cite{yao2007early}, batch normalization~\cite{ioffe2015batch}, dropout~\cite{srivastava2014dropout}, and weight decay~\cite{krogh1991simple}. This aims to push learning to a subspace of a hypothesis with manageable complexity and reduce overfitting~\cite{zhang2021understanding}. Nonetheless, researchers have found that increasing model complexity not only allows for perfect interpolation but also results in low generalization error. Various studies have been conducted to analyze such overparameterized models, i.e., trainable parameters are much higher in number than the training data points. From the statistical viewpoint, the majority of overparameterized models exhibit a double-descent effect~\cite{belkin2018understand, mei2022generalization}. In fact, the generalization error follows the traditional U-shaped curve until a specific point, after which the error decreases, and reaches a global minimum in the overparameterized regime. According to the double-descent phenomenon, the minimum generalization error tends to appear at infinite complexity, i.e., the more overparameterized the model, the smaller the error. From the optimization viewpoint, the good generalization behavior of highly overparameterized models is also commonly attributed to the inductive bias of gradient-based algorithms which helps with selecting models that generalize well despite the non-convexity, e.g.,~\cite{bartlett2021deep}. Intuitively, the large number of hidden units here represent all possible features, and hence the optimization problem involves just picking the right features that will minimize the training loss. This suggests that as we over-parametrize the networks, the optimization algorithms need to do less work in tuning the weights of the hidden units to find the right solution.

A disadvantage of overparameterized deep learning architectures is that they are highly susceptible to perturbations in adversarial or non-adversarial settings, compared to conventional, less sophisticated models. To illustrate the inherent brittleness of overparameterized neural networks, we refer to the notion of “neuron coverage” that is inspired by the code coverage in traditional software testing~\cite{deepxplore}. It involves generating synthetic test input data to trigger the neurons that have not been activated by the original test data. The success of this coverage criterion suggests that only a subset of the parameters is responsible for capturing the patterns needed for the task. The rest of the parameters might be unoptimized (almost stalled at initial random weights) or have received fewer updates over the training. The presence of these suboptimal subnetworks might not harm the model's performance under i.i.d. conditions. However, it does affect the model's robustness negatively. Any unusual changes in inputs that accidentally activate these neurons could lead to unpredictable model behaviors. In addition, any attacker can exploit the larger space of these suboptimal neurons by designing a malicious input that yields a particular model’s output.
\subsection{Underspecified ML Pipeline: One Pipeline, Many Models}
To solve an ML problem, we expect the model to encode some essential structure of the underlying distribution, which is inferred from the data using a designed ML pipeline, and is often what makes a model credible. Nevertheless, many explorations of the failures of ML pipelines that optimize for i.i.d. generalization, reveal a conflict between i.i.d. generalization and encoding credible inductive biases. It is called structural failure mode, as it is often diagnosed as a misalignment between the predictor learned by ERM and the causal structure of the desired predictor~\cite{arjovsky2019invariant}. In medical applications of ML, training inputs often include markers representing a doctor's diagnostic judgments~\cite{oakden2020hidden}. An analysis of a CNN model for diagnosing skin lesions showed that it relied heavily on surgical ink markings around skin lesions that doctors had deemed cancerous~\cite{winkler2019association}. In these situations, a predictor with credible inductive biases cannot achieve optimal i.i.d. generalization in the training distribution, because there are so-called “spurious” features that  are strongly associated with the label in the training data, but are not associated with the label in practically important settings. In fact, an iid-optimal predictor would incorporate the ink markings as they are highly correlated with positive cases, but these markings would not be expected to be present in deployment, where the predictor would itself be part of the workflow for making a diagnostic judgment. There is clearly an underspecification problem as there is not enough information (a lack of positivity) in the training distribution to distinguish between credible inductive biases and spurious relationships. Geirhos et al.~\cite{shortcutLearning} connect this underspecification issue to the notion of “shortcut learning”. They point out that there may be many predictors that generalize well in i.i.d. settings, but only a few of them align with the intended solution to the learning problem. Shortcut learning resembles surface learning of students in classrooms, relying on simple decision rules to pass an exam~\cite{shortcutLearning}. The problem with shortcuts is that they might go unnoticed during the i.i.d. testing and only occur in deployment scenarios. In the absence of large, diverse datasets, their risk is higher. For example, diagnostic data for rare or novel diseases is usually limited to small datasets, and unbiased validation data could be difficult to acquire.

Modern ML pipelines are poorly set up for satisfying the system requirements~\cite{underspecificationMLmodel}. Their i.i.d. evaluation procedure often results in multiple models with equivalent (similar) predictive risk (performance) while they encode substantially different inductive biases. This implies that the ML pipeline could not distinguish between these iid-optimal models despite their potential differences in terms of robustness. The ML pipelines must be specified and designed in a way that promotes the selection of the model, encoding credible inductive biases, to bridge the gap between testing behavior and deployment behavior.
\section{Robustness Assessment}
\subsection{Non-adversarial Shifts: From Synthetic to Real-World}
\label{non_adv_shifts}
Non-adversarial shifts, including natural data corruptions and perturbations, are rarely characterized. Only a few works in Computer Vision focus on such a type of model robustness assessment. They have introduced various benchmarking datasets~\cite{carvalho2019machine} to investigate the impact of naturally-shifted inputs on modern ML models. These benchmarks involve the introduction of corruptions and/or perturbations to standard (clean) datasets, allowing the assessment of different models' resilience against corrupted images. For instance, ImageNet-C~\cite{hendrycks2019benchmarking} and ImageNet-P~\cite{hendrycks2019benchmarking} serve as synthetic benchmarks, each focusing on distinct aspects of robustness: corruption and perturbation. These benchmarks decouple robustness benchmarking by applying image transformations to the original images from the ImageNet dataset. Corruptions involve significant changes in image statistics, offering a testing ground for out-of-distribution scenarios. The benchmark~\cite{hendrycks2019benchmarking} includes 15 types of corruption transformations selected from noise, blur, weather, and digital categories, with five severity levels controlling the degree of distribution shift. The following are examples of algorithmically-generated corruptions:
\begin{itemize}
  \item[-] \textit{Gaussian noise} simulates noise in low-lighting conditions.
  \item[-] \textit{Shot (Poisson) noise} represents fluctuations in pixel intensity.
  \item[-] \textit{Impulse noise} resembles sporadic bright or dark pixels.
  \item[-] \textit{Defocus blur} mimics the effect of an out-of-focus image.
  \item[-] \textit{Frosted Glass Blur} creates a soft, diffused effect.
  \item[-] \textit{Motion blur} represents blurring caused by rapid camera movement.
  \item[-] \textit{Zoom blur} occurs during rapid camera zooming.
  \item[-] \textit{Snow} visually obstructs details in an image.
  \item[-] \textit{Frost} creates a frosty or icy appearance.
  \item[-] \textit{Fog} creates a shrouding effect using the diamond-square algorithm.
  \item[-] \textit{Brightness} mimics changes in illumination intensity.
  \item[-] \textit{Contrast levels} vary depending on lighting conditions and object color.
  \item[-] \textit{Pixelation} occurs when upsampling a low-resolution image.
  \item[-] \textit{JPEG compression} introduces compression artifacts in images.
\end{itemize}
Indeed, each corruption varies in severity to reflect different levels of image degradation. For instance, speckle noise is an additive noise with severity controlled by the intensity of noise added to each pixel. Lower severity levels introduce minimal noise, while higher levels result in more noticeable distortions and reduced clarity. Similarly, Gaussian blur severity depends on the amount of blurring applied, determined by the size and weights of the filter. Lower severity levels apply mild blur, preserving most details, while higher levels lead to more pronounced blurring and loss of sharpness. 

Perturbations, on the other hand, entail subtle transformations of original images, drawn from the same categories as corruptions but are designed to be more challenging to perceive visually. The perturbation benchmark~\cite{hendrycks2019benchmarking} aims to assess models' performance in the face of subtle data distribution shifts. Creating a comprehensive real-world robustness benchmark that incorporates systematic distribution shifts poses challenges compared to synthetic benchmarks. The complexity arises from the multifaceted variations that can simultaneously occur in real-world data. For instance, images featuring the same object may be captured with different cameras, from diverse viewpoints, in various locations, and under different weather conditions~\cite{recht2019imagenet, hendrycks2021many}. Temporal Perturbations~\cite{shankar2021image} are natural perturbations that are deduced from small changes occurring in assembled sets of contiguous video frames that appear perceptually similar to humans, but might produce inconsistent predictions for ML models. All these perturbations result in a condition where the distribution of the test set differs from the one of the training set.

Non-adversarial or natural perturbations serve as synthetic or newly created testing datasets for evaluating models in an out-of-distribution manner, simulating shifts from their original training distribution. These perturbations effectively highlight the models' fragility to naturally occurring data distortions. However, the absence of performance degradation on these datasets does not necessarily imply robustness, as they may suffer from selection and sample bias. For instance, when creating datasets with data augmentation techniques like changing image brightness, the use of brightness-based corruptions might not be indicative of model robustness, as the training data distribution already encompasses various brightness levels. In this case, non-adversarial robustness represents how well the model maintains its predictive performance under foreseeable input shifts (i.e, variation of brightness level). 
\subsection{Adversarial Attacks: Categories and Aims}
The predominant evaluation methods for ML model robustness are the adversarial attacks, where the community invents many ways to carefully craft perturbations that can deceive a given ML model. The purpose of such attacks is to produce an adversarial example (AX), i.e., an input \( x' \) close to a valid input \( x \) according to some distance metric (i.e., similarity) or admissibility criteria (i.e., semantically-preserving modification ranges), whose model's predictions, denoted as \( f(x') \) and \( f(x) \), respectively, are different (\( f(x') \neq f(x) \)). First, they can be categorized into white-box attacks and black-box attacks according to how much knowledge an attacker has about the subject model. White-box attacks are implemented with direct access to the model or its training data, whereas black-box attacks can only access the target model through queries: pairs of inputs and outputs. The growth of black-box attacks started with the discovery of the transferability of adversarial examples~\cite{papernot2016transferability}. Cross-model transferable, where attackers can construct adversarial examples in known deep learning models and subsequently attack a related unknown model. Cross training-set transferable refers to the attacks that exploit shared vulnerabilities across different datasets or domains. Second, adversarial attacks can be targeted or non-targeted. If we consider the example of image classification, targeted attacks aim to force the classifier to output a particular (chosen) class, whereas untargeted attacks attempt to make it return any class other than the original label. This categorization of adversarial attacks is based on the attacker’s goal and information access, which determines the relevance of an attack for an application. For instance, targeted white-box attacks are more suitable for security concerns, while untargeted black-box attacks can be appropriate for assessing robustness to noise.  
\subsubsection{White-box Adversarial Attacks}
White-box adversarial attacks are a type of attack on ML models where the attacker has complete access to the architecture, parameters, and gradients of the target model. In this scenario, the attacker can exploit this knowledge on the model’s intrinsics to craft adversarial examples that are specifically designed to manipulate the model’s outcomes. Most white-box adversarial attacks employ \textbf{Gradient-based approaches}, where the attacker computes gradients of the model's loss function with respect to the input data and uses this information to guide the perturbation process. The most notorious of these approaches is the Fast Gradient Sign Method (FGSM)~\cite{FGSM}. It operates as a one-step method by computing the gradient of the model's cost function ($J$) with respect to the input data ($x$) and then perturbing the input along this gradient direction. Specifically, it alters the input ($x$) by adding noise ($\eta$) in the direction that maximizes the loss, using a magnitude defined by an epsilon value ($\epsilon$) to limit the perturbation within a certain range. The attack aims to maximize the loss by perturbing the input data without exceeding the epsilon-boundary, calculated using a distance metric (e.g., $L_\infty$ or $L_2$ norm). The formulation can be represented as:

$$x_{\text{adv}} = x + \epsilon \cdot \text{sign}(\nabla_x J(\theta, x, y))$$
Here, $x_{\text{adv}}$ denotes the adversarial example,  $x$ is the original input, $\epsilon$ represents the magnitude of perturbation, $\nabla_x J(\theta, x, y)$ signifies the gradient of the cost function with respect to the input data, and $\theta$ represents the model's parameters. 

Despite its effectiveness in rapidly generating AXs, FGSM's limitation lies in its single-step gradient-based approach to input perturbation. To address this, iterative variants of FGSM have been proposed. IFGSM~\cite{IFGSM} repeatedly applies FGSM steps with small size and clips pixel values, ensuring they are within an $\epsilon$-neighbourhood. PGD~\cite{PGD} is another iterative method, but randomly initializes examples within the \( L_{\infty} \) norm sphere and performs random restarts, differing from IFGSM.

Aside from gradient-based methods, white-box adversarial attacks employ other techniques. \textbf{Optimization-based approaches} directly optimize perturbations to maximize the model's prediction error while ensuring imperceptibility, such as Carlini \& Wagner (C\&W)~\cite{carlini2017towards} and DeepFool~\cite{deepfool}. \textbf{Model-based approaches} rely on a surrogate model to generate adversarial examples that transfer to the target model. For instance, AdvGAN~\cite{AdvGAN} employ GANs to generate adversarial examples by training the generator to craft perturbations that deceive the target model while optimizing the discriminator to differentiate between original and perturbed instances.
\subsubsection{Black-box Adversarial Attacks}
Black-box adversarial attacks involve limited access to the target model's architecture, parameters, and gradients. Despite this, attackers can interact with the model by querying input-output pairs to craft adversarial examples that alter its behaviors. Black-box adversarial attacks usually demand a high number of queries, but they are better representative of external attacks’ simulation. 

To guide the search for adversarial perturbations through model queries, black-box adversaries may leverage local-search-based approaches~\cite{narodytska2017simple}, gradient estimation~\cite{chen2017zoo, bhagoji2018practical}, combinatorial optimization~\cite{moon2019parsimonious}, and evolutionary metaheuristics~\cite{qiu2021black} like Genetic Algorithms (GA)~\cite{alzantot2019genattack}.

Many improvements have been proposed to upgrade white-box adversarial attacks for better transferability. SmoothFool (SF)~\cite{smoothfool} is an improved version of DeepFool that produces smoother perturbations to enhance the transferability of adversarial examples, making them more effective across different models or datasets. An IFGSM boosted with momentum~\cite{dong2018boosting} improves the transferability of the produced AXs across models as its stable updates aid in navigating narrow valleys, small fluctuations, and suboptimal local extrema. 

Inspired by meta-learning, Meta Gradient Adversarial Attack (MGAA)~\cite{MGAA} is a versatile method designed to blur the boundary between white-box and black-box attack gradients. By randomly selecting models from a model zoo and iteratively conducting both white-box and black-box attacks on diverse tasks, MGAA effectively enhance the transferability of adversarial examples across models and seamlessly integrates with any existing attack method.

The revelation of numerous inputs causing incorrect predictions by ML models, despite being anticipated as accurate, has garnered significant attention~\cite{ren2020adversarial}. This discovery sparked a race to create finely-tuned adversarial attacks, aiming to generate imperceptible noise capable of manipulating deep neural networks (DNNs) to produce erroneous outputs. For more comprehensive details, we refer to the survey by Costa et al.~\cite{costa2023deep}, which delves into adversarial attacks, including methodologies and recent advancements. As the adversaries grow in popularity, criticism has surfaced regarding the practicality of these digital adversarial inputs in real-world scenarios. For example, even if an attacker possesses access to the autonomous driving car's model, applying optimized imperceptible noise to a traffic sign on the road for misleading passing cars remains impractical. Although digital adversarial attacks may be successful in lab experiments or with models accessed through APIs, they are limited in the physical world. Consequently, physical adversarial attacks have emerged to execute unrestricted input alterations, which assemble all the synthetically-generated inputs without any $l_p$ norm bounding, but preserve the semantic identity of the source input. As a result, physical AXs are effective in complex real-world scenarios.
\subsubsection{Physical Adversarial Attacks}
The first physical adversarial attack~\cite{sharif2016accessorize} was released in 2016 and it fools facial recognition systems by creating adversarial eyeglass frames designed with patterns that, when worn by individuals, cause mis-recognition of the individual. It marked an important milestone in the exploration of adversarial attacks in the physical domain. Since then, many variants of physical adversarial attacks have been explored, each highlighting different concepts and vulnerabilities.

\textbf{Physical attacks by object manipulation} include affine transformations such as translations and rotations of the images that can be applied physically to the objects. They tend to expose brittleness of visual object recognition models. More broader, guess-and-check of naturally-occurring situations~\cite{gilmer2018motivating}, like changing the object position or taking pictures from another perspective angle, are able to fool visual object recognition models.

\textbf{Patch- and texture-based attacks} involve attaching adversarial patches or textures onto physical objects to deceive the targeted model. Patch-based attacks, as demonstrated by artificial sticks on traffic signs~\cite{eykholt2018robust} or natural stickers on faces~\cite{wei2022adversarial}, exploit strategically designed patches to be stuck on the target object’s surface. Texture-based attacks, as exemplified by adversarial camouflage for vehicles\cite{wang2021dual}, manipulate the appearance of objects using adversarial textures applied to 3D models.

\textbf{Environment-driven optical attacks} manipulate maliciously the environmental factors like lighting alterations, reflections, or other optical distortions that can negatively affect the model performance. Examples include exploitation of devices like projectors~\cite{huang2022spaa}, laser emitters~\cite{duan2021adversarial}, flashlights~\cite{wang2023rfla}, as well as natural occurrences like shadows~\cite{zhong2022shadows} and reflected light~\cite{wang2023rfla}.

\textbf{Model-based physical attacks} leverage generative models to produce semantic adversarial examples. For instance, SemanticAdv~\cite{qiu2020semanticadv} leverages attribute-conditioned image editing via a generator to alter specific semantic attributes like hair color, facial expressions, or a car's position on the road. These modifications retain a realistic visual appearance and resemblance to the original image while changing specific semantic details.

Research into the causes of brittleness against adversarial attacks leads to the association-based statistical learning nature of supervised ML~\cite{buckner2020understanding, ilyas2019adversarial}. Nevertheless, their prevalence in modern ML models like deep neural networks is due to overparameterization that make them tend to exploit all patterns inherent in the data that contain predictive information, including such patterns that are inscrutable to human cognition or only associative but not causal for the prediction target~\cite{freiesleben2022intriguing}. This inherent tendency might clarify why various models trained on the same dataset can be deceived by identical adversarial examples.
\subsection{DL Software Testing: Test-Driven Model Verification}
Deep Learning (DL) software testing has emerged as a class of model verification techniques which bridges the gap between adversarial attacks and non-adversarial data shifts. As such, it differs from adversarial attacks in the sense that the methods are not used to improve a model resilience, but rather to evaluate a certain number of properties (associated with robustness) after the model was refined and/or trained. Hence, it does not act on the model but rather aims at verifying it. DL software testing renovates the conventional software testing methods, including test oracle identification, test adequacy evaluation, and test input generation, to be specialized for finding unrestricted adversarial examples that expose the target DNN brittleness. The goal is to construct systematic approaches that produce synthetic inputs, representing both major and minor naturally-occurring conditions to reveal potential incorrect behaviors. This involves optimizing the search over the space of transformed data as opposed to randomly sampling non-adversarially distorted inputs (i.e., non-adversarial shifts). Therefore, the designed test cases should cover effectively a broader set of input-output mappings than i.i.d. evaluations to complement the conventional statistical testing.
\subsubsection{Pseudo Oracle}
Classical testing methods for DNN classifiers require testing the prediction of an input against a ground truth value. Even if this is possible for labeled datasets, labeling data can be labor-intensive and costly. In particular, the absence of ground truth, or ``oracle'', is known in software engineering as the oracle problem. One way to circumvent this problem is to introduce a “pseudo” oracle to test correctness of an input.. 

The most common pseudo-oracle adopted by DL software testing is metamorphic testing, which allows finding incorrect behaviors by detecting violations of identified metamorphic relations (MRs). The MRs define data transformations to derive new synthetic inputs from the original ones while preserving the relationship between their expected outputs. The most prevalent type of MRs for metamorphic DL testing is called, semantically-preserving metamorphic relations. The latter include data transformations that retain the task-related semantics; as a result, both labels of the original input and its transformed counterpart must be equal. To meet this requirement for visual recognition models, researchers~\cite{deeptest, deephunter, deepevolution} have adapted numerous image transformations that include changing lighting conditions such as brightness and contrast, applying geometric distortions such as translation and scaling, and simulating weather conditions such as fog and rain. DeepRoad~\cite{deeproad} relies on a Generative Adversarial Network (GAN~\cite{GAN})-based method to provide realistic snowy and rainy scenes, which can hardly be distinguished from original scenes and cannot be generated by DeepTest~\cite{deeptest} using simple affine transformations. DeepRoad leveraged a recent unsupervised DNN-based method (i.e., UNIT~\cite{UNIT}) which is based on GANs and VAEs~\cite{VAE}, to perform image-to-image transformations. UNIT~\cite{UNIT} can project images from two different domains (e.g., a dry driving scene and a snowy driving scene) into a shared latent space, allowing the generative model to derive the artificial image (e.g., the snowy driving scene) from the original image (e.g., the dry driving scene). 

Differential testing~\cite{mckeeman1998differential} is also a well-established pseudo-oracle that takes the shape of N-versioning, which consists in $N$ semantically equivalent models that will be used to test an input. N-versioning is strongly related to the notion of ensemble learning, which uses the knowledge of multiple models. DeepXplore~\cite{deepxplore} applies the DT approach on DL models by comparing the behaviour of multiple implementations or models for the same task. The goal of the systematic test input generation is to increase the divergence between the models’ predictions on these test inputs. However, differential testing is not limited to semantically similar models, any semantic comparison allowing building the pseudo-oracle proxy is valid. Ma et al.~\cite{ma2021selecting} leverages the DT approach based on subspecialized models, i.e., models that are trained on sliced training data only. 
\subsubsection{Test Adequacy Criteria} 
The input space of ML problems is often large, high-dimensional, which necessitates prioritizing and selecting test cases in order to reduce the model testing workload. To increase the likelihood of fault exposure, there are several proposed adequacy criteria that estimate if the generated test cases are `adequate' enough to terminate the testing process with confidence that the DNN under test will behave properly in real-world settings. 

Neuron Coverage (NC)~\cite{deepxplore} was inspired by the code coverage used for traditional software systems. NC computes the rate of activated neurons to estimate the amount of neural network's logic explored by a set of inputs. Formally, given a set of neurons $N$, the neuron coverage of a test set $T$ of inputs was originally defined as follows.
\[ 
\mathrm{NC}(T):=\frac{\#\{n \in N \mid \operatorname{act}(n, x) \geq \tau \forall x \in T\}}{\# N}
\]
where the symbol \# refers to the cardinality of $T$ and $\operatorname{act}(n, x)$ is the activation of the neuron $n$ when the test input $x$ is fed to the network. Then, Ma et al.~\cite{deepgauge} generalized the concept of NC by proposing DeepGauge, a set of multi-granularity testing criteria for DNNs, including multi-level neuron coverage criteria that capture both major function regions as well as the boundary regions of activations. Additionally, DeepGauge provides layer-level coverage metrics such as Top-k Neuron Coverage (TKNC) and Bottom-k Neuron Coverage (BKNC), measuring the activation rates of k neurons in, respectively, hyperactive state and hypoactive state on each layer. In our previous work~\cite{deepevolution}, we defined two levels of distance-based neuron coverage as follows: (i) local-neuron coverage measures the neurons covered by a generated test input that were not covered by its corresponding original input; and (ii) global-neuron coverage counts the neurons covered by a generated test input that were not covered by all previous test inputs. ‌Furthermore, Odena and Goodfellow~\cite{tensorfuzz} propose a structural coverage for DNN that considers the positions of activated neurons in the network layers via an encoding of the neurons’ activations into a trace (i.e., a concatenated vector), then, only the entries that trigger novel activation traces would be preserved along with their traces to improve the diversity of the forthcoming test generations. Li et al.~\cite{li2019structural} show that obtained failures are pervasively distributed in the finely divided space defined by such coverage criteria; so the correlation between high structural coverage and fault-revealing capabilities (i.e., failure counts) is more likely due to the adversary-oriented search rather than the resulting enhancement of the structural coverage criteria. The global neuron coverage in DeepEvolution~\cite{deepevolution} quickly reaches a state of little or no change, while the local neuron coverage remains more effective in helping the optimization process around the original test input.

The following criteria were more focused on the behavioral deviations caused by the synthetic inputs compared to their original sources and aid in the model confidence reduction while promoting for diverse inputs.

Kim et al.~\cite{kim2018guiding} proposed Surprise Adequacy (SA) that computes the distance between the activation trace spawned by a given test input and its nearest neighbor obtained by a training data input with the same actual label. Increased SA values should lead to irregular network behaviors with a high chance of uncovering hidden errors. Deepfault~\cite{eniser2019deepfault} was developed to identify the pattern of neurons that are more present in error inducing inputs, which leads to pinpoint the suspicious neurons, i.e., neurons likely to be more responsible for incorrect DNN behaviour. Then, it enables the generation of failure inducing tests through the use of suspicious neurons’ activation gradients on correctly classified examples. DeepGini~\cite{feng2020deepgini} was designed based on a statistical perspective of DNN, which allows reducing the problem of measuring misclassification probability to the problem of measuring set impurity, which allows us to quickly identify possibly-misclassified tests. Intuitively, a test is likely to be misclassified by a DNN if the DNN outputs similar probabilities for each class. Thus, the set impurity metric yields the maximum value when DNN outputs the same probability for each class. 
\subsubsection{Systematic Test Input Generation}
The above-mentioned test adequacy criteria are not used as a plain testing metric like with traditional software, but rather as a way to incrementally generate test cases that maximize/minimize those given criteria. To achieve this, techniques such as the fuzzing process to randomly mutate samples from a dataset or greedy search, and evolutionary algorithms to evolve test inputs into more fault-revealing ones.

DeepXplore~\cite{deepxplore}, leverages first-order gradient ascent algorithms to produce the inputs, maximizing simultaneously neuron coverage and multiple DNNs’ outputs divergence ratios. DeepTest and DeepRoad~\cite{deeptest} uses a coverage-guided greedy search technique to systematically explore further within the inputs that trigger uncovered neurons, in order to efficiently produce synthetic tests that can increase neuron coverage. TensorFuzz~\cite{tensorfuzz}, DLFuzz~\cite{tensorfuzz}, and DeepHunter~\cite{deephunter} relies on a coverage-guided fuzzing process that keeps transforming the test inputs triggering uncovered DNN activations. Both aim to evaluate the DNN robustness and the performance degradation caused by quantization. Last, we find the search-based approaches specialized for DL testing, which employ gradient-free optimizers (i.e, metaheuristics) to increase the fitness of the generated test inputs (i.e., faulty-revealing ability). DiffChaser~\cite{diffchaser} implements GA to expose divergences amongst different DNNs with different arithmetic precisions. DeepEvolution~\cite{deepevolution} implements different swarm-based metaheuristic algorithms to optimize the search of prominent data transformations that are likely to derive either adversarial examples or difference-inducing inputs aiming at revising, respectively, the DNN robustness assessment.
\subsection{Empirical Methods and Their Limitations}
The common limitation of these assessment approaches is their empirical nature, as they all depend on the process of data generation or the collection of shifted data to test the model's robustness against altered inputs. They compute robustness scores on this data to infer a statistical estimation of the confidence we place in the model facing unexpected changes in its input. If these methods do not uncover failed tests, it does not guarantee the model's robustness. This prompts consideration of formal verification methods that conduct a complete exploration of a system's space given a set of properties to check, as opposed to empirically-guided verification methods. Researchers~\cite{urban2021review} have proposed the implementation of formal verification methods like ReluPlex~\cite{katz2017reluplex} and FANNETT~\cite{naseer2020fannet}, to provide guarantees on a model based on specified mathematical verification criteria. However, these methods face challenges such as combinatorial explosion due to the model's size and complexity, as well as the high dimensionality of input data, making it computationally intensive to cover all possible input variations. Hence, their utilization is often restricted to certain types of models, lagging behind the state of the art and less prevalent in modern ML applications. Another significant limitation is the formalization of the property to be verified; if a property cannot be properly formulated, it cannot be verified. This explains why most verification methods~\cite{meng2022adversarial} focus on model robustness against lp norm-bounded, as defining and expressing properties for rigorous robustness verification against natural shifts or application-specific invariances is challenging.

Empirical robustness assessment provides a practical, but valuable approach to complement the i.i.d. performance by statistical scoring of the model robustness in the face of unexpected data changes. On one extreme, adversarial examples provide worst-case perturbations with a defined expectation that models should remain performant under small, bounded perturbations. On the other extreme, naturally-occurring distortions that can produce corrupted entries under various sources of noise require a controlled severity procedure to avoid generating unrecognizable inputs, even to humans. Robustness assessment needs to better account for expected model behavior between these extremes, emphasizing the necessity to identify data changes' preconditions under which a model's performance is not expected to deteriorate. Especially for robustness to natural and non-adversarial perturbations, characterizing the type of data the model might encounter is crucial. This cannot be done algorithmically based on the analysis of the model's behaviors or in-data distribution features. It is vital to find better interfaces for domain knowledge in modern ML pipelines. Domain-aware testing methods are needed to thoroughly test models on application-specific tasks and check that performance on these tasks remains stable beyond the collected data samples used as in-distribution data. Designing stress tests~\cite{underspecificationMLmodel} well-matched to applied requirements and providing good ``coverage'' of potential failure modes is a significant challenge that requires incorporating domain knowledge. For intance,~\cite{braiek2022physics} proposed to leverage physics first principles and system-related design properties in the form of input-output sensitivities to generate invariance or directional expectations tests, i.e., the prediction should remain the same or should change direction under the introduction of input changes, respectively. Thus, the failed tests expose the inconsistencies of trained aircraft performance models with the domain-specific knowledge.

Regarding data generation methods and guidance for fault-revealing test cases, we have observed modest use of generative models, and researchers should further investigate the generative modeling approach for crafting and simulating edge cases to challenge supervised ML models. Indeed, deep generative models can discover hidden patterns within in-distribution data, and then, leverage them to perform nonlinear data transformations and semantically-sounded feature-based alterations. In fact, GANs~\cite{AdvGAN, deeproad, qiu2020semanticadv} have shown their effectiveness for image-to-image transformation, semantically-attribute change, and style transfer between images, that offer more diversity and naturalness than simple pixel-value and affine transformations. There is a need to explore more recent generative models not yet applied to model robustness verification. For instance, Generative Pre-trained Transformers (GPTs)~\cite{yenduri2023generative} can be employed to create test case generators producing application-specific stress tests for supervised NLP models that go beyond rule-based transformations~\cite{ribeiro2020beyond} like word replacement and typos injection.
\section{Robustness Amelioration}
\subsection{Data-Centric Amelioration Strategies}
\subsubsection{Data Debiasing}
Data debiasing refers to the process of countering biases present in datasets used for training machine learning models to avoid unfair or skewed predictions. Sampling bias is one of the most frequent data biases in ML applications that can be handled algorithmically. The training data samples are debiased with different techniques to mitigate the proneness of ERM to learning patterns associated with the majority, which makes it vulnerable to minority inputs, such as edge cases. 

Conventional practice is to employ either upsampling and/or downsampling to mitigate sample bias in the data. Upsampling (oversampling) increases the number of instances in the minority group by generating synthetic samples or replicating existing ones. Popular algorithms for upsampling include SMOTE (Synthetic Minority Over-sampling Technique)~\cite{chawla2002smote} and ADASYN (Adaptive Synthetic Sampling)~\cite{he2008adasyn}. Downsampling (undersampling) decreases instead the number of instances in the majority group by removing them to match the minority group's size. Using mini-batch gradient optimization, we could virtually balance the group ratio of training samples by updating gradients on batches with equal input sizes for every data group in the data distribution~\cite{shimizu2018balanced}. This balanced batching method has shown to be more effective than traditional over-sampling or under-sampling methods on unbalanced classification problems.

Resampling approaches have primarily targeted on rectifying class imbalances~\cite{more2016survey}, as opposed to biases within individual classes. To apply them to debiasing variabilities within a class, subgroups must be manually identified through annotations, which necessitates a priori knowledge of the latent structure to the data. To minimize human effort, clustering algorithms have been used to identify clusters in the input data prior to training and to inform resampling the training data into a smaller set of representative examples~\cite{nguyen2008supervised}. However, this method cannot scale to high dimensional data like images or cases, where a semantic-based distance between instances is difficult to implement because it often relies on significant pre-processing to extract features. Regarding this challenge, Amini et al.~\cite{amini2019uncovering} proposed an innovative debiasing technique to adjust the respective sampling probabilities of individual data points while training. This was accomplished by designing a variational autoencoder (VAE), specifically made for debiasing, called DB-VAE. The latter learns the underlying latent variables within the data distribution in an entirely unsupervised manner. It then allows for adaptive resampling of batches of data according to their inherent attributes during training. The adaptive resampling aims to feed the learning model batches with equally-distributed latent features. Assessed for algorithmic fairness, DB-VAE has successfully trained debiased models on facial recognition data containing racial and gender biases, and significantly improves classification accuracy and decreases categorical bias.
\subsubsection{Data Augmentation}
Data augmentation aims to enhance the diversity among the training samples and enlarge the size of distinctive instances through the ERM. It involves data transformations that derive new synthetic inputs supporting the model generalization by promoting the learning of feature representations insensitive to irrelevant noises or naturally-occuring distortions. For instance, basic image augmentations can be pixel-value transformations like brightness and contrast modifiers, or geometric transformations such as random rotation, translation, and mirroring. The severity of such distortions should be tuned to prevent significantly altering the data distribution but ensures a beneficial regularization effect on the model. Other augmentation techniques~\cite{cutout, lee2020smoothmix} systematically eliminate information from the training inputs to increase the robustness of the model against occlusions and information loss, in general. The cutout augmentation~\cite{cutout} works similarly to dropout, by nullifying activations of input neurons associated with arbitrary input image patches. As a result, the network learns not to heavily rely on any specific feature of the data. In order to step beyond the null patches, Cutmix~\cite{yun2019cutmix} propose to instead inject patches that have been cut out from other images in the training dataset. In a higher level of sophistication, we can find the mixup augmentation~\cite{zhang2017mixup} that relies on linear mixing of training images and labels to encourage more stable predictions on data outside the training distribution. Mixup-derived~\cite{cutout} methods have contributed to continued enhancements in corruption robustness. PuzzleMix~\cite{kim2020puzzle} improves upon Mixup by optimizing the amount of information retained in the original pairs of images while preserving saliency information during the mixing procedure. To alleviate sharp edge effects, Smoothmix~\cite{lee2020smoothmix} creates a smooth blending mask by randomly sampling mask shape and associated shape parameters (e.g., a square mask with sampled dimensions), then, the pairs of images are combined using the relative proportions of each image derived from the mask. 

To create a complex augmentation, multiple simple augmentations can be chained together~\cite{costante2015exploring}. AugMix~\cite{hendrycks2019augmix} improves corruption robustness of models by using a weighted combination of augmentations while enforcing consistent embeddings of the augmented images. Even though studies~\cite{cutout,dodge2017can,lee2020smoothmix} emphasize the importance of combining multiple augmentations to achieve good generalization, it is a non-trivial task to find the optimal combination. In some cases, inappropriate combinations can cause a degradation in model generalization such as in ~\cite{dodge2017can} for the DeepFashion Remixed benchmark dataset. Most software model testing approaches~\cite{deepxplore,deeptest, feng2020deepgini} implement test input prioritization and systematic test generation that uses data transformations similar to augmentation techniques, but searches for more diverse transformed data with fault-revealing abilities. The generated synthetic datasets, especially those resulting in false predictions, can be added to augment the training data. Indeed, their associated labels can be deduced based on the predefined metamorphic relationship of the input-output transformation~\cite{deeptest, feng2020deepgini}, or based on the majority voting~\cite{deepxplore} to automatically generate labels for the generated test inputs. Thus, no manual labeling is required to perform a conventional supervised learning of the model using the original dataset plus a selection of synthetically-produced inputs. This retraining often results in fixing the majority of the revealed erroneous behaviors while preserving or hopefully improving its performance.

However, data augmentation has been criticized because it heavily relies on rule-based data transformations, which can lead to semantically non-meaningful combinations that are not compatible with real-world data~\cite{taori2020measuring}, e.g., a horse with a patch covering the head of a frog. Therefore, generative-based augmentations are an important initiative to produce uniquely synthetic images using generative models trained on the in-data distribution, but they can override the inherent characteristics of samples in order to cover underrepresented input regions. For instance, a style transfer variant of generative adversarial networks (GAN)~\cite{GAN} proposed in~\cite{taori2020measuring} has been shown to facilitate robustness against out-of-distribution when used as an augmentation technique~\cite{hendrycks2021many}. GAN-based solutions are also proven useful in enhancing the model's immunity to adversarial attacks~\cite{abdelaty2021gadot,wang2019direct,sun2019enhancing}. In particular, they are employed to generate adversarial samples~\cite{abdelaty2021gadot}, perturbations~\cite{wang2019direct}, and boundary samples~\cite{sun2019enhancing} to defend the networks against adversarial attacks. 
\subsection{Model-Centric Optimization Methods}
\subsubsection{Transfer Learning}
Transfer learning is a ML technique that involves transferring knowledge learned by the model in one task to improve another, related task. Transfer learning is based on the notion that DNN learns feature representations gradually from simple, task-agnostic features (e.g., lines) to complex, task-specific features (e.g., nose or ears for face recognition task) that can be fully or partially transferred to another problem. A common way to implement transfer learning is to pretrain a DNN on a large and diverse dataset and then use the first n pretrained layers (i.e., known as feature extractor) as an initialization for a new DNN that is then trained on a new dataset. Transfer learning methods are many, and the selected method determines the efficiency of the knowledge transfer from the pretrained model to the new one. In regards to robustness, it has been demonstrated that in certain settings it can increase the models OOD robustness, both in adversarial and non-adversarial settings~\cite{hendrycks2019using}. Intuitively, the use of a small supervised dataset with an overparameterized neural network may provide high ID performance, but there is no guarantee regarding its behavior against out of distribution inputs due to shortcut learning and under-fitted neurons. This risk of shortcut learning and under-fitted neurons can be diminished by using pretrained models. Furthermore, the advances in self-supervised learning (SSL) for representation learning enables the exploitation of massive unlabeled target datasets to train models that solve an auxiliary, pretext task to learn rich feature representations, e.g., learning to distinguish between images~\cite{chen2020simple,caron2021emerging}. Transfer learning with SSL~\cite{sun2019unsupervised, sun2020test} involves using these self-learned models as pretrained feature extractors.
\subsubsection{Adversarial Training}
Adversarial training~\cite{FGSM} has emerged as a class of techniques which bridges the gap between pure data augmentation and the model optimization process. In adversarial training (AT), the model parameters are iteratively adjusted to minimize the worst-case adversarial loss by incorporating adversarial examples into the training process. At its core, AT consists of alternating between two steps: 
\begin{enumerate}
    \item \textit{Finding worst-case adversarial perturbations:} For each original training input, AT computes the perturbation that maximizes the adversarial loss within a certain constraint (such as a bounded perturbation). This can be formulated as follows.
    \[
    \delta^* = \underset{\delta \in \Delta}{\operatorname{argmax}} \ell\left(h_\theta(x + \delta), y\right)
    \]
    where, $\delta$ is the adversarial perturbation added to input $x$, $\Delta$ is the set of possible perturbations, $h_\theta$ is the neural network with parameters $\theta$, $x$ is the input data, $\ell$ is the loss function, and $y$ is the true label of $x$.
    \item \textit{Training on these adversarial examples:} It takes a gradient step at these worst-case adversarial examples to update the model's parameters. Essentially, this can be stated as follows. 
    \[
    \theta := \theta - \nabla_\theta \ell\left(h_\theta\left(x + \delta^*\right), y\right)
    \]
    where, $\theta$ is the parameter vector, $\nabla_\theta$ is the gradient with respect to $\theta$, $\ell$ is the loss function, $h_\theta$ is the neural network with parameters $\theta$, $x$ is the input data, $\delta^*$ is the optimal adversarial perturbation, and $y$ is the true label of $x$.

\end{enumerate} 
Thus, AT solves a min-max problem where the inner maximization aims to find effective perturbations $\delta$ from some distribution $\Delta$ (e.g., adversarial or noise) while the outer minimization aims to update the model parameters $\theta$ to reduce expected error. 

Kurakin et al.~\cite{IFGSM} demonstrates that AT can be applied in massive datasets like ImageNet, showing an increase in robustness for one-step adversarial methods. As a result, several adversarial training approaches have been proposed to maximize the robustness of a model during deployment by adjusting the decision boundaries in response to adversarial perturbations computed based on training data.

AT with one-step attacks is prone to overfitting, especially with large datasets. The overfitting problem hinders the convergence to a global minimum, which makes the adversarially-trained models vulnerable to black-box attacks. Tramer et al.~\cite{tramer2017ensemble} leverage multiple pre-trained models to produce adversarial examples for AT. They aim to remove the tight coupling between the on-training model and the adversarial attack, which could discredit the adversarial examples in the long run. Mummadi et al.~\cite{mummadi2019defending} propose to compute the gradients w.r.t a batch of training inputs rather than one single input, then, the gradients are merged to create a shared perturbation for the entire batch of inputs. Once these shared perturbations have been iteratively refined and clipped, they are used to augment training data during AT.

The high cost of generating strong adversarial examples via gradients makes adversarial training impractical on large-scale problems like ImageNet. Shafahi et al.~\cite{shafahi2019adversarial} develop a cost-effective AT version that eliminates the overhead cost of generating adversarial examples by recycling the gradient information computed when updating model parameters. Indeed, both model parameters and image perturbations are updated simultaneously during the backward pass, rather than using separate gradient computations. This intuitive sharing of gradients has the same computational cost as non-adversarial training and can be 3-30 times faster than conventional adversarial training. 

In most cases, AT often leads to unfounded increases in the margin along decision boundaries, which negatively affects the original accuracy. To achieve a better trade-off between original and robust accuracy, several mitigation strategies have been proposed. Wang et al.~\cite{wang2019improving} suggested the differentiation between the misclassified and correctly classified original inputs during adversarial training because the minimization step is more important for misclassified examples than the maximization one that is negligible. Zhang et al.~\cite{zhang2020attacks} proposed to selectively consider the least adversarial examples (referred to as “friendly”) among the adversarial data that is confidently misclassified. These “friendly” adversarial examples can be systematically identified by adding early-stopping criteria to iterative attack methods. Zhang et al.~\cite{zhang2020geometry} added weights for data points used in AT based on their potential to derive adversarial examples. Close to the class boundary, the original data points are more susceptible to being turned into adversarial examples, which is why larger weights are assigned to them. Rade et al.~\cite{rade2021helper} found that the AT induces an unexpected increase in the margin along certain adversarial directions, thereby degrading the original accuracy. Hence, they proposed to reduce this effect by incorporating additional incorrectly-labelled data during training, which indeed improves accuracy without compromising robustness. To find out more details, we recommend Costa et al.'s survey~\cite{costa2023deep}, which elaborates on the methodologies and advances of adversarial training. 
\subsubsection{Randomized Smoothing}
Randomized smoothing~\cite{liu2018towards} is a probabilistic adversarial defense technique that involves the systematic injection of controlled random noise to original data points, which adjusts the model to become noise-invariant. Unlike adversarial training that responds to individual adversarial examples, randomized smoothing aims for a more globally resilient model through smoothing its decision boundaries. Mathematically, randomized smoothing can be represented as follows:

Given a classification model \(f\) and an input \(x\), the smoothed prediction \(S_f(x)\) is obtained by aggregating predictions over multiple noisy perturbations of the input:

\[S_f(x) = \text{argmax}_c \sum_{i=1}^{N} \mathbb{I}[f(x + \epsilon_i) = c]\]
Where:
\begin{itemize}
    \item[-] \(S_f(x)\) represents the smoothed prediction for input \(x\) under model \(f\).
    \item[-] \(N\) is the number of noisy samples or perturbations applied to \(x\).
    \item[-] \(f(x + \epsilon_i)\) denotes the model's prediction on the perturbed input \(x + \epsilon_i\) (where \(\epsilon_i\) is a random noise vector).
    \item[-] \(\text{argmax}_c\) finds the class with the maximum aggregated prediction.
\end{itemize}
The advantages of randomized smoothing~\cite{cohen2019certified} are: (i) It provides theoretical guarantees for robustness of models within certain probabilistic bounds, which is defined as certified robustness that guarantees a stable prediction for any input within a certain range. Thus, it offers a more principled approach to ensuring model robustness compared to empirical adversarial training strategies. (ii) It allows leveraging sophisticated pre-trained models as foundation to build robust ones through smoothing their decision boundaries using relevant datasets. Fast Adversarial Training~\cite{chen2022efficient} combines the strengths of randomized smoothing and single-step adversarial training to make the most of both approaches. On one hand, Fast AT applies randomized smoothing for better optimization of the inner maximization problem. On the other hand, Fast AT proposes backward smoothing initialization over single-step random initialization, which enhances smoothing in the -perturbation ball.
\subsubsection{Adapted Loss and Regularizers}
Concerning robustness, a variety of loss functions have been used to incorporate specific objectives: triplet loss~\cite{mao2019metric}, minimising distance between true and false classes because adversarial attacks shift the internal representation towards the “false” class, consistency across data augmentation strategies~\cite{tack2022consistency}, and adding maximal class separation constraints~\cite{mustafa2020deeply}. In a similar manner, adapted regularisation can be designed to make models more robust. Li and Zhang~\cite{li2021improved} propose a PAC-Bayesian approach to tackle the memorization of training labels in fine-tuning. Chan et al.~\cite{chan2019jacobian} develop an approach to optimize the saliency of classifiers’ Jacobian by adversarially regularizing the model’s Jacobian in line with natural training inputs. Moosavi et al.~\cite{moosavi2019robustness} studied the impact of AT on the curvature of the loss surface with respect to inputs. They found that AT reduced the loss curvature of the network, resulting in a more linear behavior. Hence, they proposed curvature regularization (CURE), which is a new regularizer that directly minimizes curvature of the loss surface to promote quasi-linear behavior in the vicinity of data points. CURE leads to adversarial robustness that is comparable to that of adversarial training. Furthermore, Wu et al.~\cite{wu2020adversarial} found a correlation between robust generalization and the flatness of weight loss landscapes resulting from the AT. Thus, they proposed Adversarial Weight Perturbation (AWP), which regularizes weight loss landscapes by adversarially perturbing both inputs and weights. By injecting systematically worst-case weight perturbations, AWP can be easily incorporated into any existing adversarial training approaches with little overhead. 
\subsubsection{Defensive Distillation}
Another defense class of techniques against adversarial examples is Defensive Distillation~\cite{papernot2016distillation}, which trains a Distilled Network on the predictions from a previously trained neural network (the Initial Network). By replacing dataset labels with continuous prediction values, the Distilled Network is likely to be more resilient to adversarial attacks. In light of the promising results achieved by defensive distillation, Papernot and McDaniel~\cite{papernot2017extending} suggest combining the original label with the Initial model uncertainty when training the distilled network. Zi et al.~\cite{zi2021revisiting} proposed to train robust small student DNNs using the predictions produced by an adversarially-trained larger teacher DNN as robust soft labels. As a result, the student DNN is guided to learn both natural and adversarial examples in all loss terms. Low Temperature Distillation (LTD)~\cite{chen2021ltd} uses a relatively low temperature in the teacher model and different, but fixed, temperatures for the teacher and student models. Indeed, the temperature is a scaling parameter used to adjust the sharpness of the probability distribution over classes. Lower temperatures make the distribution sharper, emphasizing the most probable class, while higher temperatures result in a softer, more uniform distribution over classes. LTD lowers the temperature of the teacher model used for crafting soft labels to boost the student model's robustness by only lowering the temperature of the teacher model for crafting soft labels without encountering the gradient masking problem (occurs when the loss gradients w.r.t inputs significantly diminish during the training of the distilled model). 
\subsection{Post-training Model Enhancements}
\subsubsection{Ensembling}  
Ensembling is a powerful technique in ML that combines multiple individual models, independently optimized on the same datasets, to achieve more accurate predictions and enhanced predictive performance compared to using a single model. The impact of ensemble learning on robustness has also been studied, and ensemble models have been shown to be more robust than a single model~\cite{grefenstette2018strength}, since redundancy can provide extra resilience. Mani et al.~\cite{mani2019towards} take another step, by training each model of an ensemble to be resilient to a different adversarial attack by injecting a small subset of adversarial examples, which profit to the ensemble globally, even though it comes at the cost of training more models.
\subsubsection{Pruning} 
Pruning is the act of removing weights connections from pre-trained NNs, which has mainly been used for model compression to reduce their footprints for on-edge deployment. Tong el al.~\cite{jian2022pruning} showed that pruning can result in models that are more resilient to adversarial attacks. The pruning is motivated by the “Lottery Ticket Hypothesis”, which basically assumes that any randomly initialized DNN contains a subnetwork that can match the same level of its accuracy when trained separately. This “winner” subnetwork will be more robust if redundant or suboptimal neurons are deleted, since erroneous behaviors will be less frequent. Studies~\cite{cosentino2019search,yan2019robust} tested this hypothesis and confirms that the pruned model tends to be more robust to adversarial attacks. Chen et al.~\cite{chen2022can} show that replacing unstable and insignificant neurons (i.e., operating in the flat area of ReLU activation) by linear function significantly raises the robustness at minimal predictive performance cost.
   
Several mechanisms have been suggested for dealing with unnecessary and/or unstable computation units by systematically identifying and trimming them. DeepCloak~\cite{gao2017deepcloak} enables the detection and elimination of unnecessary features in deep neural networks, thereby reducing the ability of attackers to craft adversarial examples. HYDRA~\cite{sehwag2020hydra} implements a DNN pruning technique that explicitly includes a robust training objective to guide the search for connections to prune. The outcome of such robustness-aware pruning is compressed models that are state-of-the-art in standard and robust accuracy. Dynamic Network Rewiring (DNR)~\cite{kundu2021dnr} is another pruning technique that defines a unified, constrained optimization formulation that combines model compression targets with robust adversarial training. In addition, DNR relies on a one-shot training strategy that achieves an overall target pruning ratio with only a single training iteration. 
\subsubsection{Model Repairs}
The strategies, which involve changing training data or updating the model, require retraining or fine-tuning, which is costly since modern neural networks are complex. Moreover, pretrained models can sometimes be obtained from a third party or training data may be confidential, making access to the whole in-data distribution challenging. Importantly, there is a disconnection between the assessment techniques and the improvement processes for model robustness. There is no guarantee that the new optimized or upgraded model will respond correctly to the counterexamples discovered during the assessment using most amelioration techniques. Robust models are often evaluated on adversarial examples or shifted inputs generated by assessment methods. It is important to show that the amelioration strategy does not overfit on counterexamples, but it is often overlooked how successful it is at fixing the revealed erroneous behaviors. 

Therefore, researchers have proposed post-training, model-level repair of DNNs, i.e., repair through the modification of the weights of an already trained model. In conventional software systems, these narrowly-defined modifications are called patches. When it comes to deep learning, the patch aims to fix the model’s erroneous behaviors against a specific subset of counterexamples. Although the high complexity of modern ML models makes it difficult to perform provable model repairs (i.e., the counterexample provided will always be corrected), heuristic repair strategies that do not guarantee to repair all observed counterexamples still have benefits when compared to retraining or fine-tuning the model. Indeed, the original neural network likely went through training and functions properly on most inputs, so applying small, less impactful but precise changes to fix brittleness makes sense. CAusality-based REpair (CARE)~\cite{sun2022causality} identifies the “‘guilty” neurons (i.e., the ones that caused the false prediction) using causality-based fault localization, then, modifies the weights of these identified neurons to reduce the misbehaviors. Arachne~\cite{sohn2023arachne} is similar to CARE, but ensures no disturbance of the model's correct behaviors. Indeed, it uses differential evolution to generate effective patches for the localised weights that can fix specific mispredictions of a DNN without drastically reducing its clean accuracy. GenMuNN~\cite{wu2022genmunn} ranks the weights based on their influences on the model’s predictions. It then generates mutants using the computed ranks and evolves them using genetic algorithms to increase the chances of finding mutants that satisfy the stopping criteria. By tracking the training history, NeuRecover~\cite{tokui2022neurecover} finds weights that have changed significantly over time. A weight becomes a subject for repair when it is no longer contributing to correct predictions in the earlier stage of training but is leading to incorrectly-predicted inputs. Similarly to NeuRecover, I-Repair~\cite{henriksen2022repairing} modifies localised weights to influence predictions for a specific set of fault-revealing inputs while minimising the impact on correct predictions. NNrepair~\cite{NNRepair} uses fault localisation to identify suspicious weights, and then uses constraint solving to modify them marginally. There are provable model repairs like PRDNN~\cite{sotoudeh2021provable}, REASSURE~\cite{fu2021sound}, and  Minimal Modifications of DNNs (MMDNN)~\cite{goldberger2020minimal}. However, these approaches are not scalable to large DNNs, only support ReLU activation, and often do not support polytope repair or multilayer repairs. 
      
% \begin{figure}
% \title{}% Figure title
% \caption{}% Figure caption
% \label{}%
% \end{figure}

% \begin{table}
% \title{}% Table title
% \caption{}% Table caption
% \label{}%
% \end{table}

\subsection{Bridging the Theoretical-Practical Divide}
Regarding adversarial robustness~\cite{silva2020opportunities}, while methods like randomized smoothing aim to provide guarantees within specific scenarios or against certain types of attacks, achieving a universally certified model that is completely immune to all possible adversarial perturbations remains an ongoing research challenge. In terms of non-adversarial robustness~\cite{drenkow2021systematic}, no method has demonstrated constant or near-constant performance across all corruptions, alterations, and severities. Few or no guarantees on performance were provided for non-adversarial distortions.

Improvements in model robustness under narrow sets of data changes, mostly synthetically generated, tend to result in narrow robustness for similar or close conditions. This indicates that while modern high-capacity models benefit from controlling covariates during training to learn robust inductive bias, the set of covariates is still small compared to the combinatorial explosion of covering all possibilities. In the field of computer vision, recent work~\cite{geirhos2021partial} suggests that increasing dataset size is a successful strategy for improving deep neural network (DNN) robustness. However, even with datasets of hundreds of millions of images, a measurable gap remains, raising sustainability concerns for this strategy.

Generative models show promise in natural, application-specific test case generation~\cite{deeproad}, leveraging semantic latent features for model debiasing~\cite{amini2019uncovering}. These models can capture how important features vary across different segments or instances of the data. They learn to transform instances from segment to segment through style transfer or generate unique examples with semantic attributes~\cite{qiu2020semanticadv}. Incorporating domain knowledge and subject-matter expertise to explain latent features discovered by generative models can lead to insights into the structural weaknesses of a model's inductive bias. 

Human-in-the-Loop ML Pipelines (HIL ML)~\cite{vaughan2018making} traditionally address learning frameworks that accommodate noisy crowd labels, known as ``learning from crowds''. Recent developments in HIL ML focus on building improved model pipelines by engaging the crowd, identifying weak components of a system~\cite{nushi2017human}, recognizing noise and biases in training data~\cite{yang2019scalpel}, and proposing data-based explanations for incorrect predictions~\cite{cabrera2021discovering}. Research studies~\cite{nanda2020unifying, ning2021improving, peterson2019human} demonstrate that training more robust models involves leveraging human uncertainty on sample labels, integrating human rationales for instance labeling into the training process, or actively querying relevant perturbations from an expert during training. While these are promising research directions, further enhancements can be achieved by exploiting deep active learning~\cite{ren2021survey}, lifelong or curriculum learning~\cite{wang2021survey}, offering alternative tactics for increasing training data efficiency. Indeed, continual learning strategies~\cite{hadsell2020embracing} are capable of enhancing the model adaptability to long-tail events and changing conditions.

The widespread use of ML in real-world applications requires an urgent transition from theoretical concepts to practical implementations of ML robustness. Nowadays, ML systems are crafted by practitioners, so their robustness enhancements depend on the practitioners’ participation, as exemplified by Jin et al.~\cite{jin2020bert}, where potential adversarial examples are collected through a sequence of engineering steps.~\cite{braiek2022physics} leverage a physics-guided adversarial testing method to craft inputs on which the model violates physics-grounded sensitivity rules that are derived beforehand by aircraft engineers. Afterward, all revealed counter-examples are exploited to perform a physics-informed regularization, which constrains the model optimization with the desired level of consistency w.r.t the physics domain knowledge. 

In practice, model robustness should be evaluated within a predefined input domain. As a result, practitioners should be provided with tools to gauge their robustness with respect to their application-specific conditions, then integrate a reject option~\cite{hendrickx2021machine} when deploying ML systems in production. This option can, for instance, rely on uncertainty or out-of-distribution layers, to effectively discard untrusted predictions. Thus, the decision about entries can then be deferred to an alternative backup treatment, e.g., involving human agents when necessary~\cite{hendrickx2021machine}.

\section{Conclusion}
The comprehensive exploration of ML robustness in this chapter culminates in the recognition of model robustness as a pivotal element in ensuring the trustworthiness and reliability of AI systems. The academic discussion traverses multifaceted dimensions, highlighting the dichotomy between adversarial and non-adversarial aspects. This distinction underscores the diverse array of unforeseen data changes that ML models face after deployment, from intentionally-crafted adversarial attacks that aim to exploit model vulnerabilities, to the subtler but equally impactful non-adversarial shifts in data distributions. These insights emphasize the necessity for robustness in safeguarding against a spectrum of potential threats and uncertainties in real-world applications.

Central to the chapter's analysis are the fundamental challenges to a truly robust ML model. It delves into the ramifications of data bias, model complexity, and the critical issue of underspecification in ML pipelines, illustrating how these factors can significantly impede the robustness of AI systems. The exploration continues through different perspectives in the domain of robustness assessment methods. Adversarial attacks, both digital and physical, serve as powerful tools to expose vulnerabilities in models, especially when confronted with maliciously crafted inputs or specific use cases. Non-adversarial shifts, including natural data corruptions and systematic distribution changes, are scrutinized in the context of real-world data challenges. Subsequently, we delve into how DL software testing leverages established techniques to enhance the search for data changes that reveal model’s faults, ultimately aiming to efficiently verify the model's robustness in the face of naturally occurring distribution shifts.

In addressing the strategies to fortify ML robustness, the chapter does not present a singular solution but rather a constellation of approaches, each with its unique strengths and limitations, offering a nuanced understanding of their potential impact and the trade-offs they entail. Techniques such as data debiasing and data augmentation are applied during data training preparation, while transfer learning, adversarial training, and adapted regularizers form the foundations of robust model optimization. Last but not least are post-training methods such as pruning and model repairs that operate directly on the original model to trim its weak components or adapt them to mitigate identified brittleness.

The conclusion of this chapter is not an end but a beginning – a call to action for continued research and innovation in the field of ML robustness. It recognizes that robustness is not a static goal but a continuous pursuit, one that requires persistent refinement and adaptation in the face of increasingly-complex models and ever-changing real-world conditions. The insights and methodologies discussed herein, not only expands the understanding of robustness in the context of ML but also emphasizes the need for a shift from theoretical concepts to practical implementations of ML robustness. It advocates providing practitioners with tools that assist them in continuously evaluating and enhancing the robustness of their models on a human-in-the-loop basis, enabling them to effectively satisfy the requirements of real-world machine learning applications, especially safety-critical ones.

  %\include{chapter2}

  %\appendix
  %\include{appendix01}
  %\include{appendix02}

%\Backmatter
%
%%%%%%%%% REFERENCES %%%%%%%%%
% 
%        ====  Option 1 ====
%        Plain LaTeX entries
%        ===================
% 
%\include{references}
%
%
%         ==== Option 2 =====
%         bibtex (+natbib)
%         ===================
%
% Elsevier's Standard Reference Styles
% https://booksite.elsevier.com/9780081019375/content/Elsevier%20Standard%20Reference%20Styles.pdf
%
%
% Model 1. Numbered style.
% %% set \citestyle{elsarticle-num} in document preamble.
\bibliographystyle{elsarticle-num}
%
% Model 2. Harvard (name-date) style.
% %% set \citestyle{elsarticle-harv} in document preamble.
%\bibliographystyle{elsarticle-harv}
%
% Model 3. Vancouver (numbered) style.
% %% set \citestyle{vancouver} in document preamble.
% \bibliographystyle{vancouver}
%
% Model 4. Embellished Vancouver (superscript numbered) style.
% %% set \citestyle{vancouver-super} in document preamble.
% \bibliographystyle{vancouver-super}
%
% Model 5. APA (American Psychological Association). Name-date style.
% %% set \citestyle{apa} in document preamble.
% \bibliographystyle{apa}
%
% Model 6. AMA (American Medical Association). Numbered style.
% %% set \citestyle{ama} in document preamble.
% \bibliographystyle{ama}
%
% Model 7. Saunders (name-date) style.
% %% set \citestyle{saunders} in document preamble.
% \bibliographystyle{saunders}
%
% Model 8. Saunders (numbered) style.
% %% set \citestyle{saunders-num} in document preamble.
% \bibliographystyle{saunders-num}
%
% Model 9. ACS (numbered) style.
% %% set \citestyle{acs-num} in document preamble.
% \bibliographystyle{acs-num}
%
% Model 9a. ACS (superscript numbered) style.
% %% set \citestyle{acs-super} in document preamble.
% \bibliographystyle{acs-super}
%
% Model 9b. ACS (name-date) style.
% %% set \citestyle{acs} in document preamble.
% \bibliographystyle{acs}
%
% Your own bibstyle.
% \bibliographystyle{<your-bib-style>}
%
% Bibdata (your-bibtex-file.bib)
\bibliography{references.bib}
%
%
%
%
%           ==== Option 3 =====
%           biblatex
%           ===================
% Remove 'natbib' package from document preamble and
% load 'biblatex' package instead.
% No biblatex styles are provided.
%
% \printbibliography
% 
\end{document}